\documentclass[11pt,a4paper]{article}
\usepackage[hyperref]{acl2020}
\usepackage{times}
\usepackage{latexsym}

\aclfinalcopy 
\usepackage{microtype}
\usepackage{amsmath}
\usepackage{booktabs}
\usepackage{color,soul}
\usepackage{amsmath}
\usepackage{graphicx}
\usepackage{url}
\usepackage{todonotes}
\usepackage{lmodern}
\usepackage{multirow}
\usepackage{array}
\newcolumntype{"}{@{\hskip\tabcolsep\vrule width 1pt\hskip\tabcolsep}}
\usepackage{dblfloatfix}
\DeclareMathOperator{\attn}

\newcommand{\RNum}[1]{\uppercase\expandafter{\romannumeral #1\relax}}

\title{Towards Automatic Generation of Questions from \textit{Long} Answers}

\author{Shlok Kumar Mishra$^{1}$, Pranav Goel$^{1}$, Abhishek Sharma$^{3}$,\\ \textbf{Abhyuday Jagannatha}$^{2}$, \textbf{David Jacobs}$^{1}$, \textbf{Hal Daum\'e \RNum{3}}$^{1,4}$  
\\
$^1$ Department of Computer Science, University of Maryland College Park 
\\
$^2$College of Information and Computer Sciences,University of Massachusetts Amherst 
\\
$^3$Axogyan AI, 
$^4$Microsoft Research 
\\
\texttt{
\{shlokm, pgoel1,djacobs,hal\}@cs.umd.edu}\\
\{abhyuday\}@cs.umass.edu
\{abhisharayiya\}@gmail.com
}

\date{}

\begin{document}
\maketitle
\begin{abstract}

Automatic question generation (AQG) has broad applicability in domains such as tutoring systems, conversational agents, healthcare literacy, and information retrieval. Existing efforts at AQG have been limited to short answer lengths of up to two or three sentences. However, several real-world applications require question generation from answers that span several sentences. Therefore, we propose a novel evaluation benchmark to assess the performance of existing AQG systems for long-text answers. We leverage the large-scale open-source Google Natural Questions dataset to create the aforementioned long-answer AQG benchmark. We empirically demonstrate that the performance of existing AQG methods significantly degrades as the length of the answer increases.  Transformer-based methods outperform other existing AQG methods on long answers in terms of automatic as well as human evaluation. However, we still observe degradation in the performance of our best performing models with increasing sentence length, suggesting that long answer QA is a challenging benchmark task for future research. 



\end{abstract}

\begin{table*}[!t]
\centering
\scalebox{0.88}{
\begin{tabular}{p{2cm} p{8.1cm} p{6.5cm}}
\toprule
\begin{tabular}[c]{@{}c@{}}Difference\end{tabular} &  \begin{tabular}[c]{@{}c@{}}Different QG Settings\end{tabular} & \begin{tabular}[c]{@{}c@{}}Examples of Previous Work\end{tabular}                                                                         \\ \midrule[\heavyrulewidth]
Type of QG system   & \begin{tabular}[c]{@{}c@{}}1. Rule-based: Heuristics for generation,\\ ML to rank results (optional).\\ \\ 2. \textbf{Neural QG}: End-to-end trained\\ neural networks.\end{tabular} & \begin{tabular}[c]{@{}c@{}}\cite{rus2010first,heilman2011automatic}\\ \cite{mazidi2014linguistic}, \\ \cite{labutov2015deep}\\  \cite{du2017learning, du2017identifying}\\ \cite{zhou2017neural, zhao2018paragraph} \end{tabular}                                       \\ \midrule
Answer Awareness    & \begin{tabular}[c]{@{}c@{}}1. \textbf{Answer-Aware QG}: Make explicit use of\\ given target answer span.\\2. Use of context without explicit control\\ over the specific answer span.
\end{tabular} & \begin{tabular}[c]{@{}c@{}}\cite{zhou2017neural, yuan2017machine}\\  \cite{song2018leveraging, zhao2018paragraph} \\ \\ \cite{du2017learning}\end{tabular}                                       \\ \midrule
Answer Length       & \begin{tabular}[c]{@{}c@{}}1. Short-answer setting: Target answer\\ spans \textit{1-3 sentences on average}.\\2. \textbf{Long-answer setting}: Target answer\\ spans \textit{4 or more sentences on average}.\end{tabular}                                & \begin{tabular}[c]{@{}c@{}}\cite{du2017learning, zhao2018paragraph}\\ \cite{sachan2018self} \\ (\textit{No prior work}\\ to the best of our knowledge)\end{tabular} \\ \bottomrule
\end{tabular}}
\caption{Summary of different QG settings (where source is always textual). This is not an exhaustive list of settings and previous works. Bold-faced options indicate the setting our work is situated in. Data used in all our experiments is in the English language.}
\label{tab:different_settings}
\end{table*}

\section{Introduction}

Automatic Question generation (AQG) refers to automatic generation of a natural text question for a given natural text answer. Automated QG has several application domains such as patient literacy \citep{lalor2018comprehenotes}, healthcare \citep{raghavan2018annotating,pampari2018emrqa}, education \citep{le2015evaluation,kuyten2012fully}, automated testing\citep{brown2005automatic}, and information retrieval \citep{chali2015towards}. Additionally, large-scale access to affordable and quality education has fueled the recent surge of \textbf{M}assive \textbf{O}pen \textbf{O}nline \textbf{C}ourses (\textbf{MOOC}s) that can benefit from accurate AQG systems. AQG has also been posed as a proxy task for machine comprehension \citep{yuan2017machine}. Generated questions can also serve as additional weak-labels for joint-training of question-answering and AQG systems \cite{sachan2018self} or can possibly be used in self-training schemes. Due to its wide applicability, there is a growing interest in improving automated AQG systems \citep{nguyen2016ms,dunn2017searchqa, trischler2017newsqa, kocisky2018narrativeqa} to produce \emph{semantically relevant} and \emph{natural-sounding} questions. However, the current AQG systems focus on generating questions from short-answers only (less than 2-3 sentences). 


Generating questions for long text answers is a natural extension of automated question generation, applicable to instructional domains such as, education \citep{holme2003assessment,shapley2000line,baral2007evaluation,yudkowsky2019assessment} including examination-based assessment \citep{ory1983improving, livingston2009constructed, ganzfried2018optimal} and patient literacy of health records. Especially, clinical documents often contain long dependency-spans for relevant information \citep{jagannatha2016h}.
This can result in long answers to patient-centric questions such as ``Why is the patient having this adverse reaction?''. In the context of using AQG framework for weakly-supervised learning, long answer questions have the potential to improve the modeling of large context in machine comprehension systems. Therefore, it's important to focus on developing high-performing long-answer AQG (\textbf{LAGQ}) systems in order to better model the specifications of several practical applications.

LAQG has largely been ignored due to the lack of large-scale long-text answer datasets. Most popular AQG datasets  \citet{rajpurkar2016squad,nguyen2016ms} contain short-answer questions such as. Datasets such as \citet{Yang2015WikiQAAC} have long-answer questions but, their small sample-size prohibit the use of deep-learning models . The recent work of \citet{kwiatkowski2019natural} released a large-scale Natural Questions (NQ) corpus \cite{kwiatkowski2019natural} that contains text passages from Wikipedia and user-generated questions/answer pairs based on the entire page/passage. Google NQ dataset was designed to serve as a benchmark machine comprehension task with long-text passages. We leverage the Google NQ dataset as a repository of long-answers with associated questions to design experiments that uncover the challenges of the transition from the current AQG setting to an LAQG setting. Our work establishes the first benchmark study for LAQG using an exhaustive set of existing AQG models and aims to motivate further research in this domain. 

The main challenge in LAQG stems from the requirement of modeling dependencies across a much larger context as compared to the traditional AQG systems. Therefore, we perform a comprehensive evaluation of several NLG models under LAQG setting to analyze their performance for long-answer inputs. The analyzed models are Recurrent-Neural-Network (RNN) or Transformer-based Networks with various mechanism aimed at improving NLG for questions. RNNs  have been successfully used in several AQG systems \cite{du2017learning,zhou2017neural,gao2018difficulty,subramanian2018neural,wang2018qg,zhao2018paragraph,sachan2018self}. Similarly, transformer-based networks \citep{vaswani2017attention} have been widely used in several natural language processing tasks \cite{devlin2018bert,libovicky2018input} and have shown promising results\citep{radford2019language} towards capturing long-range dependencies in text. We also investigate the use of answer summary in guiding the question generation by using a Multi-Source Transformer-based Attention mechanism (\textbf{MSTA}). The primary input to the MSTA is the long-answer itself, with a neural-network generated summary vector of the answer as the secondary input. It's in contrast with the previous approaches \cite{zhao2018paragraph,du2017learning} that use the answer-containing paragraph as additional contextual information for short answer question generation.

We use common automated NLG evaluation methods such as BLEU and also carry out human evaluation studies to evaluate the performance of the analyzed and proposed methods.


Our main contributions are-
\vspace{-1em}
\begin{itemize}
    \setlength\itemsep{-0.5em}
    \item \textbf{Long-answer QG task or \textbf{LAQG}}: We introduce the task of question generation from long-text answers, referred to as LAQG. We redesign the recently introduced Google NQ dataset as a benchmark for this task.
    \item \textbf{Empirical Study}: We provide a thorough empirical analysis of the existing AQG systems under LAQG setting, using both standard automated evaluation metrics and human evaluation.
    \item \textbf{LAQG performance of Transformer vs LSTM}: We use our study to show the different behaviours of Transformer-based and RNN-based networks when we increase the length of answers in the context of LAQG. 
    
    
\end{itemize}

\begin{table*}[t]
\centering
\scalebox{0.88}{
\begin{tabular}{p{17cm}}
\toprule
\footnotesize
\begin{tabular}[c]{@{}l@{}}\em{\textbf{Annotated Long Answer:}} \\ Succession to the British throne is determined by descent gender for people born before October legitimacy and religion. \\ Under common law the Crown is inherited by a sovereign s children or by a childless sovereign's nearest collateral line. \\ The Bill of Rights and the Act of Settlement restrict succession to the throne to the legitimate Protestant descendants \\ of Sophia of Hanover that are in communion with the Church of England. Spouses of Roman Catholics were disqualified \\ from until the law was amended in. Protestant descendants of those excluded for being Roman Catholics are eligible.\end{tabular} \\ \midrule
\begin{tabular}[c]{@{}l@{}}\em{\textbf{Original Query:}} who will take the throne after the queen dies\end{tabular} \\ \midrule
\begin{tabular}[c]{@{}l@{}}\em{\textbf{Transformer Generated Question:}} who is next in line for the throne\end{tabular} \\ \midrule
\begin{tabular}[c]{@{}l@{}}\em{\textbf{Max-Pointer Generated Question:}} who is known for the third throne of peace\end{tabular} \\
\bottomrule                               
\end{tabular}}

\caption{Example of an annotated long answer and original query from Google Natural Questions corpus, along with the question generated by our transformer model (Section \ref{sec:our_transformer_variants}) and the previous LSTM approach using maxout pointer (Section \ref{sec:baselines}). Another example shown in Table A1 in Appendix A.}
\label{Tab:NQ_examples}
\end{table*}

\section{Task Definition and Baseline Architectures} \label{sec:proposed}
In this section, we describe the problem formulation, and define the proposed approaches for the LAQG task.

\subsection{Problem Definition}
Formally, AQG refers to the NLG task of generating a natural language question, $Q$, from a natural language answer, $A$. The training objective is 
\begin{equation}
    \mathcal{L(\theta)} = \frac{1}{n}\sum_{i}^{n}log(p(q_i|a_i;\theta))
\end{equation}
with the conditional probability $p(q|a;\theta)$ for a given question $q$ and answer $a$ parameterized by $\theta$. The predicted question $\hat{q}$ for a given $a$ is obtained through MAP inference 
\begin{equation}
    \hat{q} = \underset{q}{\operatorname{argmax \: }} p_\theta(q|a).
\end{equation}

\subsection{Transformer Networks}
\label{sec:transformer}
Transformer models are based on the attention mechanism proposed in \cite{bahdanau2014neural}. They can also use multi-head attention \cite{vaswani2017attention} to model long-range dependencies required for LAQG. The input to a transformer network is first fed through a self-attention layer followed by a feed-forward network. The Self-attention layer contains ``multi-headed'' attention, where each head attends to different parts of the sequence to leverage (global) information from the complete input sequence. Since the attention mechanism does not model sequential information, positional embeddings are used to provide a notion of locality within an input sequence. 




\subsection{Multi-Source Transformer}
\label{sec:multi_source}
 Previous studies \citet{libovicky2018input} have shown the benefits of additional information for guiding machine translation systems. Additionally, long-text answers can have multiple different questions each one targeting different aspects \citet{hu2018aspect} or difficulty levels \citet{gao2018difficulty}. Therefore, we use multi-source transformer to provide extra contextual information to help guide the question generation under LAQG setting. Specifically, we propose to input the summary sentence, extracted using a reinforcement-learning based summarization approach \citet{narayan2018ranking}, as an additional signal in a two-source transformer setup (Table. \ref{fig:multi-transformer}).
\section{Experiments}
In this section, we first discuss the dataset used in our experiments in Sec.~\ref{sec:datasets} followed by the descriptions of the contemporary AQG systems and their variants in Sec.~\ref{sec:baselines} and Sec.~\ref{sec:our_transformer_variants} and some implementation details in Sec.~\ref{ssec:details}.
\subsection{Data}
\label{sec:datasets}
\textbf{Google Natural Questions (NQ)} \citet{kwiatkowski2019natural}: 
Google NQ is a Question-Answer dataset that contains real-world user questions obtained from Google search with the associated answers as human-annotated text-spans from Wikipedia. It is built using aggregated search engine queries with annotations for \textit{long answers} (typically a paragraph from the relevant Wikipedia page).\footnote{DuReader is a similar Chinese dataset \cite{he2018dureader}; we focus on English.} 
In this dataset, we select all questions with "long-answer" tag and filter out cases where answers don't start with the HTML tag paragraph, for paragraphs. This leaves 77501 examples which we split 90/10 for training/validation, and 2136 from the original dev-set, which we use as testing data. With an average length of target answers greater than 4 sentences (Table \ref{tab:data_stats}), Google NQ serves as a dataset allowing us to propose and explore the LAQG setting.\footnote{We do not use WIKI QA \cite{rosso2017two} because it only contains 3,047 questions.}

\begin{table}[t]
\scalebox{0.8}{
\begin{tabular}{ p{1.5cm}  p{2cm}  p{2.3cm} p{1.8cm}}
\toprule
\textbf{Dataset} & \textbf{\#Training  Examples} & \textbf{Average \#Sentences } & \textbf{Average \#Words } \\ \midrule
\textbf{NQ} & 77,501                                                                         & 4.59                                                                                            & 77.92                                                                                      \\
\textbf{SQuAD}                    & 70,484                                                                          & 2.19                                                                                          & 32.86                                                                                       \\
\bottomrule
\end{tabular}
}
\caption{Statistics of the "long-answer" tag filtered Google NQ and SQuAD v1.1 datasets.}
\label{tab:data_stats}
\end{table}

\subsection{Baselines}
\label{sec:baselines}

We provide a benchmarking study of multiple AQG approaches that have been shown to provide good performance under short-answer setting. 

\begin{enumerate}
    \item \textbf{\citet{du2017learning}} proposed the first end-to-end trained neural approach for QG and showed that it significantly outperformed rule-based approaches on SQuAD v1.1. 
    They used an RNN encoder-decoder \cite{bahdanau2014neural, cho2014learning} approach with global attention \cite{luong2015effective}. 
    
    \item \textbf{Copy Mechanism}: \citet{zhao2018paragraph} employ copy-pointer mechanism \citet{gu2016incorporating} to the encoder-decoder to outperform \citet{du2017learning} on SQuAD v1.1. It treats every word as a copy target, and uses the attention scores over the input sequence to calculate the final score for a word as the sum of all scores pointing to that word. Intuitively, it works because short-answer questions frequently have large overlaps with the resulting answer, which may not hold true under LAQG setting. 
    
    \item \textbf{Maxout Pointer}: \citet{zhao2018paragraph} proposed a Maxout Pointer based approach which helps reducing the repetitions that occur in generated questions (it occurs more frequently in longer sequences). It works by picking the maximum score of word probabilities instead of combining all the probability scores.  
\end{enumerate}    

\begin{table*}[h!]
  \begin{center}
    \scalebox{0.9}{
    \begin{tabular}{@{}p{1cm}cccccc@{}}
     \toprule
     & \multicolumn{1}{c}{\textbf{B1}} &
      \multicolumn{1}{c}{\textbf{B2}} &
       \multicolumn{1}{c}{\textbf{B3}} &
        \multicolumn{1}{c}{\textbf{B4}} &
     \multicolumn{1}{c}{\textbf{M}} & \multicolumn{1}{c}{\textbf{R}} \\ \midrule[\heavyrulewidth]
    \multicolumn{1}{c}{\textbf{\cite{du2017learning}}} & \multicolumn{1}{c}{29.37} & \multicolumn{1}{c}{19.32} & \multicolumn{1}{c}{14.55}& \multicolumn{1}{c}{9.97} & \multicolumn{1}{c}{14.31} & \multicolumn{1}{c}{28.25}\\  \midrule
    \multicolumn{1}{c}{{Copy Mechanism  \citep{gu2016incorporating}}} & \multicolumn{1}{c}{{33.81}} & \multicolumn{1}{c}{{22.73}} &\multicolumn{1}{c}{16.41}
    &\multicolumn{1}{c}{12.07}
    &\multicolumn{1}{c}{16.62}
     &\multicolumn{1}{c}{33.99}\\ \midrule
    \multicolumn{1}{c}{{Maxout pointer\citep{zhao2018paragraph}}} & \multicolumn{1}{c}{34.07} & \multicolumn{1}{c}{23.18} &\multicolumn{1}{c}{{16.78}}
    &\multicolumn{1}{c}{{12.37}}
    &\multicolumn{1}{c}{{16.68}}
     &\multicolumn{1}{c}{{34.26}}\\ \midrule\midrule
    \multicolumn{1}{c}{{Transformer \citep{vaswani2017attention}}} &
    \multicolumn{1}{c}{{36.37}} & \multicolumn{1}{c}{{25.09}} & \multicolumn{1}{c}{18.27} &
    \multicolumn{1}{c}{{13.35}} & \multicolumn{1}{c}{{17.32}} &
    \multicolumn{1}{c}{34.13}\\ \midrule
    \multicolumn{1}{c}{{Transformer $+$ Copy \citep{vaswani2017attention}}} & \multicolumn{1}{c}{{34.74}} & \multicolumn{1}{c}{{23.54}} &\multicolumn{1}{c}{16.97}
    &\multicolumn{1}{c}{12.38}
    &\multicolumn{1}{c}{17.01}
     &\multicolumn{1}{c}{24.64}\\ \midrule
     \multicolumn{1}{c}{{Transformer\_iwslt\_de\_en \citep{Ott2018ScalingNM}}} &  \multicolumn{1}{c}{\textbf{36.80}} & \multicolumn{1}{c}{\textbf{{25.58}}} &\multicolumn{1}{c}{\textbf{18.78}}
    &\multicolumn{1}{c}{\textbf{13.90}}
    &\multicolumn{1}{c}{\textbf{17.45}}
     &\multicolumn{1}{c}{\textbf{35.56}}\\ \midrule
     \multicolumn{1}{c}{{Transformer\_wmt\_en\_fr\_big \citep{Ott2018ScalingNM}}} &  \multicolumn{1}{c}{{34.48}} & \multicolumn{1}{c}{{23.52}} &\multicolumn{1}{c}{16.98}
    &\multicolumn{1}{c}{12.41}
    &\multicolumn{1}{c}{16.25}
     &\multicolumn{1}{c}{33.89}\\ \midrule
     \multicolumn{1}{c}{{Transformer\_wmt\_en\_de\_big \citep{Ott2018ScalingNM}}} &  \multicolumn{1}{c}{{34.79}} & \multicolumn{1}{c}{{23.43}} &\multicolumn{1}{c}{16.79}
    &\multicolumn{1}{c}{12.16}
    &\multicolumn{1}{c}{16.30}
     &\multicolumn{1}{c}{33.92}\\ \midrule
     
     
     \midrule
      \multicolumn{1}{c}{{Multi-Source Transformer (Summary)}} & \multicolumn{1}{c}{{36.04}} & \multicolumn{1}{c}{{24.67}} &\multicolumn{1}{c}{18.00}& \multicolumn{1}{c}{{13.30}} & \multicolumn{1}{c}{{16.80}} &\multicolumn{1}{c}{34.58}\\ \midrule
      \multicolumn{1}{c}{{Multi-Source Transformer (First Sentence)}} & \multicolumn{1}{c}{{36.01}} & \multicolumn{1}{c}{{24.71}} &\multicolumn{1}{c}{18.00}
      & \multicolumn{1}{c}{{13.27}} & \multicolumn{1}{c}{{16.97}} &\multicolumn{1}{c}{35.02}\\ \midrule 
      \multicolumn{1}{c}{{ Transformer Summary Baseline}} &  \multicolumn{1}{c}{{33.79}} & \multicolumn{1}{c}{{22.70}} &\multicolumn{1}{c}{16.34}
    &\multicolumn{1}{c}{11.89}
    &\multicolumn{1}{c}{15.70}
     &\multicolumn{1}{c}{32.58}\\ \midrule
     \multicolumn{1}{c}{{Transformer First sentence Baseline}} & \multicolumn{1}{c}{{33.66}} & \multicolumn{1}{c}{{22.34}} &\multicolumn{1}{c}{16.08}&
     \multicolumn{1}{c}{{11.81}} & \multicolumn{1}{c}{{15.52}} &\multicolumn{1}{c}{32.17}\\ \bottomrule
    \end{tabular}} 
    \caption{Performance in terms of automatically computed metrics (B: BLEU4, M: METEOR, R: ROUGE) for LAQG on the Google NQ. The first three baselines use an RNN/LSTM based encoder-decoder approach (Section \ref{sec:baselines}). Transformer-based variants, row 4--8, show significant improvements over RNN/LSTM approaches, specially the Transformer\_iwlst variant (Section \ref{sec:our_transformer_variants}). Lastly, Multi-Source Transformer models, row 9--10, don't show improvement over single source transformer, which uses only the answer. The last two rows show the results for Transformer-model when fed with the summary or the first sentence as the input.}
    \label{tab:results}
  \end{center}
\end{table*}

\subsection{Proposed Transformer-based Variants for LAQG}
\label{sec:our_transformer_variants}
This section builds upon the basic Transformer model, explained in Sections ~\ref{sec:transformer} and \ref{sec:multi_source}, and proposes different variants for LAQG task. 

Transformer models based on \citet{vaswani2017attention} (Section \ref{sec:transformer}): 
\begin{itemize}
    \item\textbf{Transformer $+$ Copy:} We seek motivation from the benefits of copy-mechanism shown in \citet{zhao2018paragraph} and propose to add copy-mechanism to transformer. We do it by using the final attention output from transformer as the attention scores to guide the copy-mechanism. 
    
    \item\textbf{Transformer\_iwslt\_de\_en model:}  Transformer\_iwslt\_de\_en is the best performing model on English to German (En-De) on IWSLT'14 training data\cite{Ott2018ScalingNM}. This model has 6 encoder and decoder layers, 16 encoder and decoder attention heads, 1024 embedding dimension, and embedding dimension of feed forward network(ffn)  is 4096. 
    
    \item\textbf{Transformer\_wmt\_en\_de\_big\cite{Ott2018ScalingNM}} Transformer\_wmt\_en\_de\_big is best performing model on English to German (En-De) on WMT'16 training data\cite{Ott2018ScalingNM}. Transformer\_wmt\_en\_de\_big  has 6 encoder and decoder layers, 4 encoder and decoder attention heads, 1024 embedding dimension, and embedding dimension of ffn is 1024. 
    \item\textbf{Transformer\_wmt\_en\_fr\_big\cite{Ott2018ScalingNM}}
    Transformer\_wmt\_en\_fr\_big is best performing model on English to French (En-Fr) on WMT'16 training data\cite{Ott2018ScalingNM}. 
    Transformer\_wmt\_en\_fr\_big  has 16 encoder and decoder layers,16 encoder and decoder attention heads, 1024 embedding dimension, and ffn embedding dimension of 4096.

    
\end{itemize}
Multi-Source Transformers based on \citet{libovicky2018input} (Section 
\ref{sec:multi_source}): 
\begin{itemize}
    \item\textbf{Multi-Source Transformer (Summary)}: We extract summary sentence  from the answer-text using a reinforcement learning based approach \citep{narayan2018ranking}. We generate questions using summary sentence as the input to the multi-source transformer model (along with the target answer), to assess the benefits of summary sentence as an additional signal. 
    
    \item\textbf{Multi-Source Transformer (First Sentence)}: We also experimented with using just the first sentence of the long target answer as the additional input instead of the answer summary. 
\end{itemize}

\subsection{Implementation Details of Transformer} \label{ssec:details}
We have implemented our models on top of OpenNMT-py \cite{opennmt} and Fairseq toolkits.\footnote{https://github.com/pytorch/fairseq} We have used 5 Encoder Layers and 5 Decoder layers. Our final tuning parameters are- $lr =  0.0005, dropout = 0.3, \beta_1 = 0.9, \beta_2 = 0.98.$ using Adam \cite{kingma2014adam}.

\section{Results and Discussion}
\label{sec:results}
This section discusses the results of our models on Google NQ dataset under the proposed LAQG setting. 
\subsection{Comparison using Standard Automatic Metrics}
\label{sec:result_standard}
Table \ref{tab:results} shows the LAQG results for all the models (Section \ref{sec:baselines}) on NQ. 
We observe that the Transformer\_iwslt\_de\_en \citep{Ott2018ScalingNM} model with 6 encoder and decoder layers, 16 attention heads is the best performing model. The best performing RNN based model is the Maxout pointer model.

In MSTA models, Multi-source Transformer(Summary) performs better than Multi-source Transformer(First sentence). We also observe that Transformer models outperform all MSTA models. A possible reason for this could be that long answer passages may have multiple themes or aspects. Therefore, one long answer passage may correspond to multiple different questions depending on the aspects under consideration. A discrepancy between the aspect of a secondary input to MSTA and theme of its corresponding gold question can result in reduced automated metric evaluation.  

We believe that a guided QG approach using Multi-source Transformer has merits and can be useful for focusing question generation towards a particular aspect of the long answer. 

Finally, we also design baseline experiments to evaluate the complexity introduced by the inclusion of long answers in the question generation task. To evaluate this, we use short summary sentences generated by the model from \citet{narayan2018ranking} as inputs to a single source transformer. This model is then trained to generate the corresponding question in NQ dataset. Our experiment aims to convert the long answer QG task to a short answer QG by compressing the answer to a short sentence summary. Table \ref{tab:results} shows that the performance of Transformer model trained on a summary sentence is significantly lower than the same model trained on long answers. This suggests that LAQG cannot simply be solved by compressing the long answer. Replacing the short summary sentence by a short first sentence results in similar trends.




\begin{table*}[!t]
\begin{center}
\scalebox{0.9}{
 \begin{tabular}{c p{2.5cm} p{2.2cm} p{2.5cm} p{2.5cm} }
 \toprule
 \#Sentences & Fluency (Transformer) & Fluency (Maxout LSTM) & Correctness (Transformer) & Correctness (Maxout LSTM)\\ [0.5ex] 
 \midrule[\heavyrulewidth]
   1 & 3.71 & 3.58 & 2.55 & 2.59\\
   \midrule
   2 & 3.63 & 3.61 & 2.46 & 2.49\\
 \midrule
   3 & 3.70 & 4.09 & 2.46 & 2.59\\
   \midrule
   4 & 4.03 & 3.87 & 2.69 & 2.41\\
 \midrule
    5 & 4.68 & 2.72 & 3.74 & 1.76\\
 \midrule
 6 & 4.46 & 2.54 & 3.22 & 1.64\\
 \bottomrule
\end{tabular}}
\caption{Human Evaluation Results: Transformer model does better than the previous LSTM based QG approach in terms of grammatical fluency as well as correctness of the QA pair. This gap in human ratings increases with increasing number of sentences in the target answer.}
\label{tab:human_evaluation_results}
\end{center}
\end{table*}
\subsection{Human Evaluation}
Due to the high cost of human evaluation, we restrict our human evaluation to the best performing RNN  (maxout pointer in Section \ref{sec:baselines}) and Transformer baseline  (Transformer\_iwslt in Section \ref{sec:our_transformer_variants}).  
We use two metrics for evaluation, namely, ``fluency'' and ``correctness''.  \textbf{Fluency}, defined in the guidelines by \citep{du2017learning,stent2005evaluating}, measures the grammatical coherence  of the generated questions. \textbf{Correctness} judges whether the provided answer input is a correct answer to the generated question. We randomly selected 100 question answer pairs and asked 6 English speakers to rate the generated questions on above mentioned metrics across all the pairs. We use at least two annotators per pair. Our annotations have an average inter-annotator agreement of 0.36 for fluency and 0.42 for correctness. We group the generated questions by the sentence length of their corresponding answers. Table \ref{tab:human_evaluation_results} shows that the questions generated by the transformer model are rated higher in terms of fluency and correctness by humans. The gap between the performance of the two models increases with the increasing number of sentence lengths. Our results suggest that the Transformer architecture can more effectively model long-term dependencies over the text and are better suited for LAQG task.


\begin{table}[!t]
\begin{center}
\scalebox{0.9}{
 \begin{tabular}{c p{2.5cm} p{2.2cm} p{2.5cm} p{2.5cm} }
 \toprule
 \#Words & Transformer & (Maxout LSTM)\\ [0.5ex] 
 \midrule[\heavyrulewidth]
   0-50 & 14.10 & 12.59\\
   \midrule
    50-100 & 13.96 & 12.34\\
 \midrule
    100- & 13.93 & 12.37\\
 \bottomrule
\end{tabular}}
\caption{Our best-performing transformer model (Transformer\_iwslt\_de\_en in Table \ref{tab:results}) vs LSTM with maxout pointer on LAQG when the length of answer is in different 50-word bins.}
\label{fig:multi-transformer}
\end{center}
\end{table}

\section{Other Analysis}
This section discusses the results of increasing the answer length and analysis of Multi-Source and Single-Source Transformers.

\label{sec:analysis}

\subsection{Transformer vs LSTM with Increasing Answer Length}
\label{sec:analysis_length}
We also analyse the performance of our models on varying sentence lengths using automated NLG evaluation metrics. We observe a positive correlation between model performance degradation and answer length increase for RNN-based maxout pointer baseline as well as our Transformer model. However, our best Transformer model consistently outperforms the best RNN models across all answer lengths (Table \ref{fig:multi-transformer}). A more fine-grained version of the Table \ref{fig:multi-transformer} is included in the Appendix A1 (Tables Appendix A2  and Appendix A3). These fine-grained results also support the finding that Transformer LAQG models are more robust to increased answer length. 

\begin{table*}[!t]
\centering
\scalebox{0.88}{
\begin{tabular}{p{17cm}}
\toprule
\footnotesize
\begin{tabular}[c]{@{}l@{}}\em{\textbf{Annotated Long Answer:}}\\ The Kingdom of England and the Kingdom of Scotland fought dozens of battles with each other. They fought typically \\ over land particularly Berwick, Upon  Tweed  and the Anglo Scottish border frequently changed as a result.  Prior to the \\ establishment of the two kingdoms in the 10th and 9th centuries  their predecessors the Northumbrians  and the Picts or \\ Dal Riatans also fought a number of battles.  Major conflicts between the two parties include the Wars of Scottish \\ Independence(1296 --1357) and the Rough Wooing(1544--1551) as well as numerous smaller campaigns and individual \\ confrontations.  In 1603 England and Scotland were joined in a  personal union  when King James VI of Scotland \\ succeeded to the throne of England as King James I. War between the two states largely ceased  although the Wars of the \\ Three Kingdoms in the 17th century  and the Jacobite Risings of the 18th century  are sometimes characterised as Anglo \\  Scottish conflicts  despite really being British civil wars.\end{tabular} 
\\ \midrule
\footnotesize
\begin{tabular}[c]{@{}l@{}}\em{\textbf{Original Query:}}Why did scotland go to war with england?\end{tabular}
\\ \midrule
\footnotesize
\begin{tabular}[c]{@{}l@{}}\em{\textbf{Generated Summary}} The Kingdom of England and the Kingdom of Scotland fought dozens of battles with each other.\end{tabular}
\\ \midrule
\end{tabular}
}
\caption{Example of Summary sentence generated using the RL based approach in \citet{narayan2018ranking}. The generated summary sentence does not capture the information that is needed to answer the question which could explain lack of improvement in automatic metric scores for Multi-Source Transformer (Summary) (Table \ref{tab:results}).}
\label{Tab:example_summary}
\end{table*}

\subsection{Multi-Source vs Single-Source Transformers for LAQG}
\label{sec:analysis_multisource}
MSTA uses a sentence as a secondary input to guide the question generation. Therefore, it can be expected to perform better than single-source transformer variants if the secondary input is relevant and correct. However, we note that our MSTA models do not exhibit an increase in performance as compared to single source Transformer models (Table \ref{tab:results}). A possible explanation for this may be that our secondary input can focus on an aspect that is different from the one used by the gold question. Since our summary extraction model \citep{narayan2018ranking} is trained on a different text corpus, it may frequently produce noisy summary sentences on our QG data, or may focus on an aspect of the passage that does not align well with the gold answer. Similarly, the first sentence baseline uses the first sentence of the answer as the secondary input Which may not be relevant to the gold question.  An example of the discrepancy between gold question and the summary sentence can be seen in Table \ref{Tab:example_summary}.

\section{Related Work} 
\label{sec:rel_work}

AQG is an important problem in natural language generation (NLG) and has received significant attention in recent years \citet{pan2019recent}. 
Broadly, AQG can be divided into two categories: \textit{a) heuristic rules-based}, where questions are created using pre-defined rules or from manually constructed templates followed by ranking \citep{Heilman2010GoodQS,Mazidi2014LinguisticCI,labutov2015deep}, and \textit{2. learning-based}. 
Rule-based approaches are difficult to scale to multiple domains and often perform poorly, and therefore, they gave way to the more versatile learning-based approaches. 
Several learning-based AQG approaches are modelled as neural sequence-to-sequence (seq2seq) models \citep{du2017learning,song2018leveraging,zhao2018paragraph,yuan2017machine,zhou2017neural} based on the encoder-decoder framework. 


\citet{yuan2017machine} uses supervised and reinforcement learning with various rewards and policies to generate better quality questions. 
\citet{song2018leveraging} show the benefit of modeling the shared information between the (short) answer and the paragraph containing answer. 
\citet{kim2019improving} on the other hand, report that over-relying on the context can degrade the performance and propose to separate the answer from the context by masking it with a special token. These and other methods used reading comprehension datasets with short answers (one to three sentences long on average) either as spans within longer paragraphs such as SQuAD \citep{rajpurkar2016squad} and NewsQA \citep{trischler2017newsqa} or human-generated short answers in MS MARCO \citep{nguyen2016ms} and Narrative QA \citep{kocisky2018narrativeqa}. 
Other datasets include multiple-choice questions in RACE \citep{lai2017race} and documents (but no answers) to generate questions from in LearningQ \citep{chen2018learningq}, both of which focused on the education domain. 

To the best of our knowledge, we are the first to tackle the task of question generation for long answers using the Google Natural Questions corpus \citep{kwiatkowski2019natural} with answers spanning 4 or more sentences on average (Table \ref{tab:data_stats}). 
A non-exhaustive list of relevant works divided according to various types of AQG settings is given in Table \ref{tab:different_settings}. 

A major limitation of focusing on AQG with short answers is that the insights and the approaches that work well for short answer may not generalize to LAQG setting. 
For example, several NLG models \citep{zhao2018paragraph, Sun2018AnswerfocusedAP, Dehghani2018UniversalT} use \emph{copy-mechanism} \citep{gu2016incorporating} with seq2seq architectures that affords copying words from the input to the generated text. 
As discussed in \ref{sec:baselines} shows improvements in short answer AQG, but
unfortunately, the use of copy-mechanism for LAQG does not help, and in fact, can hurt performance in terms of both automatic and human evaluation (Tables \ref{tab:results} and \ref{tab:human_evaluation_results}). 
On the other hand, however, the use of maxout pointer approach for QG \citep{zhao2018paragraph} serves as a strong baseline for LAQG. 

Transformer-based models have now become an integral approach in the state-of-the-art systems for many NLP tasks like translation, QA, and text classification \citep{vaswani2017attention, Radford2018ImprovingLU, devlin2019bert}. 
Because such models can better handle long-range dependencies using self-attention, and can also go beyond fixed-length contexts \citep{dai2019transformer}. 
Since LAQG requires connecting the dots across a large range of text - in the form of a long answer - to generate an adequate question, we explore transformers for our task. 
We report superior performance of transformer-based models in terms of automatic and human evaluation than previous seq2seq models for AQG. 
In order to be able to provide an additional \emph{contextual} signal in addition to the long-text answer, we use multi-source transformers \citep{libovicky2018input} with multiple inputs (we use two). We show that using a summary sentence extracted from the answer or the first sentence of the answer as additional input did not improve performance, and we analyze the potential reason in Section \ref{sec:results}. 
 


\section{Conclusion}
\label{sec:conclusion}
We take the first step towards question generation in the long-answer setting, where answers can contain 4 or more sentences on average. We benchmark newly released Natural Questions corpus for question generation with both transformer networks (used for the first time in context of QG) and previously used LSTM networks. We show transformer models outperform LSTM-based models in terms of automatically computed metrics like BLEU as well as human evaluation. We also provide a detailed empirical study showing the effect of answer length on commonly used QG approaches. Our work can be directly applied to different domains such as improve the quality of question long-answer pairs in educational testing.


\bibliographystyle{acl_natbib}
\bibliography{acl2020}

\appendix

\end{document}